\documentclass[]{article}

\usepackage{graphicx}
\usepackage{amsmath}
\usepackage{amssymb}
\usepackage{booktabs}

\usepackage{multirow}

\usepackage[table,xcdraw]{xcolor}

\usepackage{bm,bbm}

\usepackage[pagebackref,breaklinks,colorlinks]{hyperref}

\usepackage[capitalize]{cleveref}
\crefname{section}{Sec.}{Secs.}
\Crefname{section}{Section}{Sections}
\Crefname{table}{Table}{Tables}
\crefname{table}{Tab.}{Tabs.}

\begin{document}

\title{SLAMs: Semantic Learning based Activation Map for Weakly Supervised Semantic Segmentation}

\author{Junliang Chen, Xiaodong Zhao, Minmin Liu, Linlin Shen\thanks{Corresponding Author}\\
Shenzhen University \\
{\tt\small \{chenjunliang2016,zhaoxiaodong2020,liuminmin2020\}@email.szu.edu.cn} \\
{\tt \small llshen@szu.edu.cn}}

\maketitle

\begin{abstract}
   Recent mainstream weakly-supervised semantic segmentation (WSSS) approaches mainly relies on image-level classification learning, which has limited representation capacity. In this paper, we propose a novel semantic learning based framework, named SLAMs (Semantic Learning based Activation Map), for WSSS.
\end{abstract}

\section{Introduction}

Semantic segmentation is one of the fundamental tasks in computer vision, which aims to assign a category to each pixel of the image. Due to the development of the fully convolutional network (FCN), many fully-supervised semantic segmentation (FSSS) methods \cite{fcn,deeplabv1,deeplabv2,bisenet,setr,segformer} have achieved excellent performance and can be widely applied. However, FSSS is built upon the accurate pixel-level segmentation labels, which can be really time consuming and requires huge labour costs. To reduce the massive resources cost of pixel-level annotation, weakly-supervised semantic segmentation (WSSS) aims at achieving comparable performance with FSSS using weaker supervision, such as bounding boxes \cite{sdi,bcm}, scribbles \cite{scribblesup,rawks}, points \cite{whatpoint}, and image-level labels \cite{aepsl,gain,cpn,ecsnet}. These methods follow a two-stage paradigm: generating pseudo semantic segmentation labels based on these weak labels, and then training an FSSS network using the generated pseudo labels. Among these annotations, image-level labels are the most conveniently acquired ones and have been widely studied. Therefore, in this work, we focus on WSSS with image-level supervision.

As the image-level annotations can not provide accurate location information, most of WSSS methods are built upon Class Avtivation Map (CAM) \cite{cam}, which relies on image-level classification learning. The original CAM is simple and effective, but has an obvious weakness, \emph{i.e.}, under-activation. It only produces high response in the discriminative regions, leading to incompletely activated object regions. To address this issue, recent approaches have attempted to expand the area of CAM \cite{mdc,seam} or adversarially erase the regions with high response and force the network to include less discriminative regions \cite{aepsl,gain,occse}. However, they still suffer from under-activation problem. Besides, they may sometimes falsely activate some background regions. The main reason for the failure of these approaches can be explained by the poor representation capacity of image-level classification learning. On one hand, the binary classification label can only tell whether a category exists or not in a given image, but fails to provide more details about where and what is the category in the given image. On the other hand, the image-level classification learning tends to build connections with the most relative regions. The most relative regions are mainly the most discriminative object regions of each category, and may include the closely related background regions which often co-occur, \emph{e.g.}, the railway in a train image.

In this paper, we make the following contributions:
\begin{itemize}
    \item We propose a novel framework, SLAMs (Semantic Learning based Activation Map) for WSSS.
\end{itemize}

\section{Related Work}
The prosperity of WSSS benefits a lot from the development of Class Activation Map (CAM) proposed by Zhou \emph{et al.} \cite{cam}. In this section, we introduce different manners to generate CAM, including image-level classification and recent approaches in new paradigm.

\subsection{Image-level Classification Learning} Most recent WSSS methods built on CAM are trained with image-level classification learning, which suffers from under-activation of the less discriminative regions. To overcoming this problem, \cite{sec,dsrg,ssdd,cian,affinitynet, seam} attempted to expand the initial CAM using boundary information or pixel-level relationship. However, they introduced additional complicated modules or training procedures. Recently, Li \emph{et al.} proposed two-stage GAIN \cite{gain} consisting of CAM generation and classification stages with a shared classifier. The second stage minimizes the classification score of the masked image for each category using the thresholded CAM in the first stage. OC-CSE \cite{occse} improved GAIN \cite{gain} by using a pre-trained and fixed classifier in the second stage. However, Zhang \emph{et al.} \cite{cpn} indicated that the training phase of these methods is unstable as they will still lose some regions more or less, due to the randomness of the hiding process. Besides, they may suffer from over-activation problem, \emph{i.e.} false activation in background regions. It results in very small classification loss in the second stage which exacerbates the training instability.

\subsection{Recent New Paradigm} Recently, some approaches explores new paradigms to generate CAM. CLIMS \cite{clims} introduces extra text knowledge with Contrastive Language-Image Pre-training (CLIP) \cite{clip} to conduct cross language image matching. With the open set knowledge available in the CLIP model trained with extra 400 million image text pairs, CLIMS can suppress the falsely activated background regions to a certain extent and thus makes the generated CAM more complete. AMN \cite{amn} proposes an activation manipulation with a per-pixel classification loss to penalize the most discriminative regions and promote the less discriminative regions using refined initial CAM, and thus generates more complete CAM. MCTFormer \cite{mctformer} introduces advanced transformer architecture to WSSS and achieves much higher performance than convolutional neural network (CNN) based approaches. For fair comparisons, we focus on CNN based approaches in this paper.

\section{Experiments}
\subsection{Dataset and Evaluation Metric}
We conduct our experiments on PASCAL VOC 2012 datset \cite{voc} with 21 categories (20 object categories and one background category), and MSCOCO \cite{coco} with 81 categories (80 object categories and one background category). For PASCAL VOC 2012, as most approaches do, we use the augmented \emph{trainaug} split (10528 images) with only image-level labels for training. The \emph{train} split (1464 images) is used to validate our method. The \emph{val} (1464 images) and \emph{test} (1456 images) split are used to evaluate our WSSS approach and compare with other methods, respectively. We report all the experimental results in the standard mean Intersection over Union (mIoU) metric for semantic segmentation. For MSCOCO, we use \emph{train} split with around 80K images for training, and \emph{val} split with around 40K images for evaluation.

\section{Conclusions}
In this paper, we propose a novel framework, Semantic Learning based Activation Map (SLAMs), for weakly supervised semantic segmentation.

{\small
\bibliographystyle{ieee_fullname}
\bibliography{references}
}

\end{document}